# Heuristic Ranking in Tightly Coupled Probabilistic Description Logics

Thomas Lukasiewicz[1]    Maria Vanina Martinez[1]    Giorgio Orsi[1,2]    Gerardo I. Simari[1]

[1] Department of Computer Science, University of Oxford, UK
[2] Institute for the Future of Computing, Oxford Martin School, University of Oxford, UK
email: firstname.lastname@cs.ox.ac.uk

## Abstract

The Semantic Web effort has steadily been gaining traction in the recent years. In particular, Web search companies are recently realizing that their products need to evolve towards having richer semantic search capabilities. Description logics (DLs) have been adopted as the formal underpinnings for Semantic Web languages used in describing ontologies. Reasoning under uncertainty has recently taken a leading role in this arena, given the nature of data found on the Web. In this paper, we present a probabilistic extension of the DL $\mathcal{EL}^{++}$ (which underlies the OWL2 EL profile) using Markov logic networks (MLNs) as probabilistic semantics. This extension is tightly coupled, meaning that probabilistic annotations in formulas can refer to objects in the ontology. We show that, even though the tightly coupled nature of our language means that many basic operations are data-intractable, we can leverage a sublanguage of MLNs that allows to rank the atomic consequences of an ontology relative to their probability values (called ranking queries) even when these values are not fully computed. We present an anytime algorithm to answer ranking queries, and provide an upper bound on the error that it incurs, as well as a criterion to decide when results are guaranteed to be correct.

## 1 Introduction

Recently, it has become apparent that Semantic Web formalisms must be able to cope with uncertainty in a principled manner. The Web contains many examples where uncertainty comes in [22]: as an inherent aspect of Web data (such as in reviews of products or services, comments in blog posts, weather forecasts, etc.), as the result of automatically processing Web data (for instance, analyzing a document's HTML Document Object Model usually involves some degree of uncertainty), and as the result of integrating information from many different heterogeneous sources (such as in aggregator sites, which allow users to query multiple sites at once to save time). Finally, inconsistency and incompleteness are also ubiquitous as the result of over- and under-specification, respectively. To be applicable to Web-sized data sets, any machinery developed for dealing with uncertainty in these settings must be scalable. In this paper, we develop an extension of $\mathcal{EL}^{++}$ [1] by means of a probabilistic semantics based on Markov logic networks [20]. $\mathcal{EL}^{++}$ is a DL that combines tractability of several key reasoning problems with enough expressive power to model a variety of ontologies; for instance, it is expressive enough to model real-world ontologies such as the well-known SNOMED CT, large segments of the Galen medical knowledge base, as well as the Gene Ontology. Moreover, $\mathcal{EL}^{++}$ underlies the OWL2 EL profile, in which basic reasoning problems are solvable in polynomial time, and highly scalable implementations are available. One of the key aspects of the extension presented here is that it is *tightly coupled*, meaning that probabilistic annotations can refer to objects in the ontology, providing greater expressive power than similar efforts in the literature.

In the area of Web data extraction [16], uncertainty comes into play even for very basic tasks. Consider, as an example, the *form-labeling* problem, consisting of the association of blocks of text (i.e., the *labels*) to the corresponding *fields* in a Web form [6], illustrated in Figure 1, where three possible instances of the problem with decreasing likelihood of occurrence are shown. In the first case, the fields $f_1$ and $f_2$ have horizontally aligned blocks of text on their left and on their right; this case exemplifies the most typical situation where the blocks of text on the left of a field (denoted $l_1$ and $l_2$) are the actual labels, while those on the right (denoted $t_1$ and $t_2$) are either unrelated or carry additional information about the fields. The second case represents the symmetric situation, where the labels are on the right of the fields. This situation is typical for eastern Web sites where the content has right-to-left reading order (e.g., for Arabic). Another possibility is exemplified by the third case, where the labels occur in the north-west region of the field. In this setting, probabilistic DL formalisms are well-suited to the task, since they couple powerful modeling capabilities with sound and complete reasoning procedures.

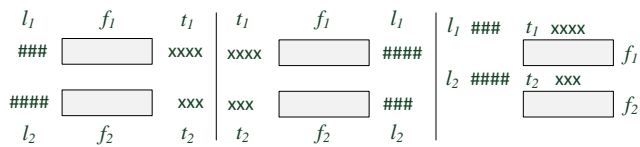

Figure 1: Uncertainty in form labeling.

**Example 1.** Consider the following $\mathcal{EL}^{++}$ formulas describing a simple ontology of Web forms. The ontology forces each field to be associated with a block of text representing its label. Fields and text blocks are disjoint sets.

$Field \sqsubseteq \exists label.Text; \quad dom(label) \sqsubseteq Field;$
$Field \sqcap Text \sqsubseteq \bot; \quad ran(label) \sqsubseteq Text.$ ∎

The set of formulas above simply defines what a correct labeling of the form should be, without providing a measure of the likelihood that the labeling is correct. Example 1 shows the need for probabilistic modeling languages for reasoning over structured data on the Web, e.g., to leverage the statistical evidence that certain patterns are more likely to occur than others.

Markov logic networks [20] (MLNs), which were developed in recent years, are a simple approach to generalizing classical logic; their relative simplicity and lack of restrictions has recently caused them to be well-received in the reasoning under uncertainty community. To our knowledge, this work is the first to develop an MLN-based probabilistic extension of DLs, and in particular of $\mathcal{EL}^{++}$.

The following are the main contributions of this paper:

(i) Introduction of tightly coupled probabilistic (TCP) DLs, an expressive class of probabilistic ontology languages that allow probabilistic annotations to refer to objects in the ontology. The probabilistic semantics is based on MLNs.

(ii) Complexity results establishing the intractability of basic computations over TCP ontologies necessary to rank atomic consequences based on their probabilities.

(iii) The proposal of *conjunctive MLNs* (cMLNs), a subset of the MLN formalism. Though we prove that this model restriction does not alleviate the complexity issues of the general case for TCP ontologies, we propose the analysis of a special kind of *equivalence classes* of possible worlds, for which we prove useful properties: deciding emptiness and generation of members can both be done in polynomial time in the data complexity, all worlds in a class are guaranteed to have the same probability, and the highest-probability classes can be identified in polynomial time in the data complexity.

(iv) An anytime algorithm for answering *ranking queries* over TCP ontologies with cMLNs that leverages the properties of equivalence classes and is guaranteed to inspect worlds in decreasing order of probability. We provide an upper bound on the error that is incurred by this algorithm based on the number of worlds and equivalence classes inspected, and a criterion to decide when results are guaranteed to be correct.

The rest of this paper is organized as follows. Section 2 describes the preliminaries on the DL $\mathcal{EL}^{++}$ and MLNs. Section 3 presents tightly coupled probabilistic DLs, and a complexity result showing the intractability of computing probabilities of atoms. In Section 4, we present conjunctive MLNs, and we analyze how their structure can be leveraged in the definition of well-behaved equivalence classes over possible worlds, as well as in an anytime algorithm that exploits this equivalence class approach in order to tractably compute a heuristic answer to the ranking of atoms according to their probabilities. Finally, Sections 5 and 6 discuss related work and conclusions, respectively.

## 2 Preliminaries

In this section, we briefly recall the description logic (DL) $\mathcal{EL}^{++}$ and Markov logic networks (MLNs).

### 2.1 The DL $\mathcal{EL}^{++}$

We now recall the syntax and the semantics of $\mathcal{EL}^{++}$, a tractable DL especially suited for representing large amounts of data. Intuitively, DLs model a domain of interest in terms of concepts and roles, which represent classes of individuals and binary relations between individuals, respectively. While we restrict ourselves to $\mathcal{EL}^{++}$ here, the general approach continues to be valid for any DL or ontology language for which instance checking is data-tractable (for instance, the closely related Datalog+/– family of ontology languages contains such tractable subsets [3]). However, note that all results in this paper were derived for $\mathcal{EL}^{++}$, and thus certain results may not hold for other logics.

**Syntax and Semantics.** We first define concepts and then knowledge bases and instance checking in $\mathcal{EL}^{++}$. We assume pairwise disjoint sets $\mathbf{A}$, $\mathbf{R}$, and $\mathbf{I}$ of *atomic concept names*, *role names*, and *individual names*, respectively. Concepts are defined inductively via the constructors shown in the first five rows of the table in Figure 2; this table adopts the usual conventions of using $C$ and $D$ to refer to concepts, $r$ to refer to a role, and $a$ and $b$ to refer to individuals. The semantics of these concepts is given, as usual in first-order logics, in terms of an *interpretation* $\mathcal{I} = (\Delta^\mathcal{I}, \cdot^\mathcal{I})$. The *domain* $\Delta^\mathcal{I}$ comprises a non-empty set of individuals and the *interpretation function* $\cdot^\mathcal{I}$ maps each concept name $A \in \mathbf{A}$ to $A^\mathcal{I} \subseteq \Delta^\mathcal{I}$, each role name $r \in \mathbf{R}$ to a binary relation $r^\mathcal{I}$ over $\Delta^\mathcal{I} \times \Delta^\mathcal{I}$, and each individual name $a \in \mathbf{I}$ to an individual $a^\mathcal{I} \in \Delta^\mathcal{I}$. The ex-tension of $\cdot^{\mathcal{I}}$ to arbitrary concept descriptions is defined via the constructors in Figure 2.

| Name | Syntax | Semantics |
|---|---|---|
| Top | $\top$ | $\Delta^{\mathcal{I}}$ |
| Bottom | $\bot$ | $\emptyset$ |
| Nominal | $\{a\}$ | $\{a^{\mathcal{I}}\}$ |
| Conjunction | $C \sqcap D$ | $C^{\mathcal{I}} \cap D^{\mathcal{I}}$ |
| Existential Restriction | $\exists r.C$ | $\{x \in \Delta^{\mathcal{I}} \mid \exists y \in \Delta^{\mathcal{I}} :$ $(x,y) \in r^{\mathcal{I}} \wedge y \in C^{\mathcal{I}}\}$ |
| Concrete Domain | $p(f_1,...,f_k)$ for $p \in \mathcal{P}^{\mathcal{D}_j}$ | $\{x \in \Delta^{\mathcal{I}} \mid \exists y_1,...,y_k \in \Delta^{\mathcal{D}_j} :$ $f_i^{\mathcal{I}}(x) = y_i, 1 \leq i \leq k$ $\wedge (y_1,...,y_k) \in p^{\mathcal{D}_j}\}$ |
| GCI | $C \sqsubseteq D$ | $C^{\mathcal{I}} \subseteq D^{\mathcal{I}}$ |
| RI | $r_1 \circ ... \circ r_k \sqsubseteq r$ | $r_1^{\mathcal{I}} \circ ... \circ r_k^{\mathcal{I}} \subseteq r^{\mathcal{I}}$ |
| Domain Restriction | $dom(r) \sqsubseteq C$ | $r^{\mathcal{I}} \subseteq C^{\mathcal{I}} \times \Delta^{\mathcal{I}}$ |
| Range Restriction | $ran(r) \sqsubseteq C$ | $r^{\mathcal{I}} \subseteq \Delta^{\mathcal{I}} \times C^{\mathcal{I}}$ |
| Concept Assertion | $C(a)$ | $a^{\mathcal{I}} \in C^{\mathcal{I}}$ |
| Role Assertion | $r(a,b)$ | $(a^{\mathcal{I}}, b^{\mathcal{I}}) \in r^{\mathcal{I}}$ |

Figure 2: Syntax and Semantics of $\mathcal{EL}^{++}$ (rep. from [1]).

Reference to concrete data objects is accomplished via the parameterization of $\mathcal{EL}^{++}$ by concrete domains $\mathcal{D}_1,...,\mathcal{D}_n$, which correspond to OWL data types. Such concrete domains are pairs $(\Delta^{\mathcal{D}}, \mathcal{P}^{\mathcal{D}})$, where $\Delta^{\mathcal{D}}$ is a set and $\mathcal{P}^{\mathcal{D}}$ is a set of predicate names; each $p \in \mathcal{P}^{\mathcal{D}}$ has an arity $n > 0$ and an extension $p^{\mathcal{D}} \in (\Delta^{\mathcal{D}})^n$. The link between the DL and concrete domains is accomplished via the introduction of *feature names* **F** and the concrete domain constructor included in Figure 2; $p$ is used to denote a predicate of a concrete domain, and $f_1,...,f_k$ to denote feature names. The interpretation function maps each feature name $f$ to a partial function from $\Delta^{\mathcal{I}}$ to $\bigcup_{1 \leq i \leq n} \Delta^{\mathcal{D}_i}$, where in general it is assumed that $\Delta^{\mathcal{D}_i} \cap \Delta^{\mathcal{D}_j} = \emptyset$ for $1 \leq i < j \leq n$.

$\mathcal{EL}^{++}$ **Knowledge Bases.** A KB consists of two sets, referred to as the ABox and the TBox, respectively containing the *extensional* knowledge about individual objects and *intensional* knowledge about the general notions for the domain in question. The ABox formally consists of a finite set of concept assertions and role assertions, while the TBox is comprised of a finite set of *constraints*, which can be general concept inclusions (GCIs), role inclusions (RIs), domain restrictions (DRs), or range restrictions (RRs) (cf. Figure 2). An interpretation $\mathcal{I}$ is a model of a TBox $\mathcal{T}$ (resp., ABox $\mathcal{A}$) iff, for each contained constraint (resp., assertion), the conditions in the "Semantics" column of Figure 2 are satisfied. We note that the expressive power of $\mathcal{EL}^{++}$ allows the expression of role hierarchies, role equivalences, transitive roles, reflexive roles, left- and right-identity rules, disjointness of complex concept descriptions, and the identity and distinctness of individuals.

Finally, here we adopt the syntactic restriction presented in [1] to avoid intractability/undecidability, which prevents intricate interplay between role inclusions and range restrictions; we do not describe it here for reasons of space.

## 2.2 Markov Logic Networks

Markov logic networks (MLNs) [20] combine first-order logic with Markov networks (abbreviated MNs; they are also known as Markov random fields) [19]. We now provide a brief introduction first to MNs, and then to MLNs.

**Markov Networks.** A Markov network (MN) is a probabilistic model that represents a joint probability distribution over a (finite) set of random variables $X = \{X_1,...,X_n\}$. Each random variable $X_i$ may take on *values* from a finite *domain* $Dom(X_i)$. A *value* for $X = \{X_1,...,X_n\}$ is a mapping $x: X \to \bigcup_{i=1}^{n} Dom(X_i)$ such that $x(X_i) \in Dom(X_i)$; the *domain* of $X$, denoted $Dom(X)$, is the set of all values for $X$. An MN is similar to a Bayesian network (BN) in that it includes a graph $G = (V, E)$ in which each node corresponds to a variable, but, differently from a BN, the graph is undirected; in an MN, two variables are connected by an edge in $G$ iff they are conditionally dependent. Furthermore, the model contains a *potential function* $\phi_i$ for each (maximal) clique in the graph; potential functions are non-negative real-valued functions of the values of the variables in each clique (called the *state* of the clique). In this work, we assume the *log-linear* representation of MNs, which involves defining a set of *features* of such states; a feature is a real-valued function of the state of a clique (we only consider binary features in this work). Given a value $x \in Dom(X)$ and a feature $f_j$ for clique $j$, the probability distribution represented by an MN is given by $P(X = x) = \frac{1}{Z} \exp\left(\sum_j w_j \cdot f_j(x)\right)$, where $j$ ranges over the set of cliques in the graph $G$, and $w_j = \log \phi_j(x_{\{j\}})$ (here, $x_{\{j\}}$ is the state of the $j$-th clique). The term $Z$ is a normalization constant to ensure that the values given by the equation above are in $[0,1]$; it is given by $Z = \sum_{x \in Dom(X)} \exp\left(\sum_j w_j \cdot f_j(x)\right)$. Probabilistic inference in MNs is intractable; however, approximate inference mechanisms, such as Markov Chain Monte Carlo, have been developed and successfully applied.

**Markov Logic Networks.** In the following, let $\Delta_{MLN}$, $\mathcal{V}_{MLN}$, and $\mathcal{R}_{MLN}$ denote the set of constants, variables, and predicate symbols. The main idea behind Markov logic networks (MLNs) is to provide a way to soften the constraints imposed by a set of classical logic formulas. Instead of considering worlds that violate some formulas to be impossible, we wish to make them less probable. An MLN is a finite set $L$ of pairs $(F_i, w_i)$, where $F_i$ is a formula in first-order logic over $\mathcal{V}_{MLN}$, $\Delta_{MLN}$, and $\mathcal{R}_{MLN}$, and $w_i$ is a real number. Such a set $L$, along with a finite set of constants $\Delta_{MLN}$, defines a Markov network $M$ that contains: (i) one binary node corresponding to each element of the Herbrand base of the formulas in $L$ (i.e., all possible ground instances of the atoms), where the node's value is 1 iff the atom is true; and (ii) one feature for every possible ground instance of a formula in $L$. The value of the feature is 1 iff the ground formula is true, and the weight of the feature is the weight corresponding to the formula in $L$. From this characterization and the description above of the graph corresponding to an MN, it follows that $M$ has an edge between any two nodes corresponding to ground atoms that appear together in at least one formula in $L$. Furthermore, the probability of $x \in Dom(X)$ given this ground MLN is

$$P(X = x) = \frac{1}{Z} \cdot \exp\left(\sum_j w_j \cdot n_j(x)\right), \quad (1)$$

where $n_i(x)$ is the *number of ground instances* of $F_i$ made true by $x$, and $Z$ is defined as above. This formula can be used in a generalized manner to compute the probability of any setting of a subset of random variables $X' \subseteq X$.

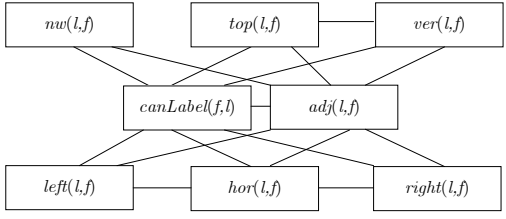

Figure 3: The graph representation of a simple grounding of the MLN from Example 2.

**Example 2.** Consider again the form-labeling problem of Example 1. The fact that certain text blocks are more likely to represent field labels than others (which is part of their *phenomenology*) can be described by assigning weights to the following first-order formulas:

$\phi_1 : canLabel(Y, X) \wedge hor(X, Y) \wedge left(X, Y) \wedge adj(X, Y);$

$\phi_2 : canLabel(Y, X) \wedge hor(X, Y) \wedge right(X, Y) \wedge adj(X, Y);$

$\phi_3 : canLabel(Y, X) \wedge ver(X, Y) \wedge top(X, Y) \wedge adj(X, Y);$

$\phi_4 : canLabel(Y, X) \wedge nw(X, Y) \wedge adj(X, Y).$

Formula $\phi_1$ describes labels that are left-adjacent and horizontally aligned with the field. The dual case, with right-adjacent labels, is captured by $\phi_2$. The case of top-adjacent labels that are vertically-aligned to the field is expressed by formula $\phi_3$, while the last expression ($\phi_4$) describes the case of labels appearing in the north-west area of the field.

The formulas above are part of the MLN associated with the $\mathcal{EL}^{++}$ formulas describing the form ontology. As an example, the following MLN formulas reflect the likelihood of the particular labeling phenomenology.

$\psi_1 : (\phi_1, 9), \quad \psi_2 : (\phi_2, 6), \quad \psi_3 : (\phi_3, 5), \quad \psi_4 : (\phi_4, 1). \quad \blacksquare$

Figure 3 shows a simple grounding of the MLN described in Example 2, relative to the set of constants $\{f, \ell\}$ (resp., one field and one label). This grounding also assumes that we have additional constraints (not shown here for simplicity) that only allow arguments of predicates to take the type of argument they expect; this sort of constraints is provided in certain implementations of MLNs such as Tuffy[1], where they are called *predicate scoping rules*.

## 3 Tightly Coupled Probabilistic DLs

Considering the basic setup from Sections 2.1 and 2.2, we now present the language of probabilistic $\mathcal{EL}^{++}$.

### 3.1 Syntax

Recall that we have (as discussed in Sections 2.1 and 2.2) an infinite universe of individual names $\mathbf{I}$, a finite set of concept names $\mathbf{C}$ and role names $\mathbf{R}$, a finite set of constants $\Delta_{MLN}$, an infinite set of variables $\mathcal{V}_{MLN}$, and a finite set of predicate names $\mathcal{R}_{MLN}$ (such that $\mathcal{R}_{MLN} \cap \mathbf{C} = \emptyset$ and $\mathcal{R}_{MLN} \cap \mathbf{R} = \emptyset$). Finally, recall that the set of random variables in the MLN coincides with the set of ground atoms over $\mathcal{R}_{MLN}$ and $\Delta_{MLN}$; alternatively, we denote this set with $X = \{X_1, \ldots, X_n\}$, as in Section 2.2.

**Substitutions and Unifiers.** We adopt the usual definitions from classical logic. A substitution is a function from variables to variables or constants. Two sets $S$ and $T$ *unify* via a substitution $\theta$ iff $\theta S = \theta T$, where $\theta A$ denotes the application of $\theta$ to all variables in all elements of $A$ (here, $\theta$ is a unifier). A *most general unifier* (mgu) is a unifier $\theta$ such that for all other unifiers $\omega$, there exists a substitution $\sigma$ such that $\omega = \sigma \circ \theta$.

**Translation into FOL.** In the rest of this work, we assume that $\mathcal{EL}^{++}$ TBoxes and ABoxes are translated into their equivalent first-order logic formulas; therefore, when clear from the context, we refer to *axioms* or *assertions* without distinguishing them from their translations into FOL. This is required for technical reasons, such as the need to be able to explicitly refer to the variables in the axioms in order to have the possibility of linking such variables with those in probabilistic annotations. Note that this does not affect the expressiveness of our formalism, nor its tractability, since the translation to FOL (and back to $\mathcal{EL}^{++}$) can be done in polynomial time in the size of the ontology.

Informally, probabilistic ontologies consist of a finite set of first-order logic formulas that correspond to the translation of $\mathcal{EL}^{++}$ axioms; each such formula is associated with a probabilistic annotation, as described next.

**Definition 1.** A *probabilistic annotation* $\lambda$ relative to an MLN $M$ defined over $\mathcal{R}_{MLN}$, $\mathcal{V}_{MLN}$, and $\Delta_{MLN}$ is a (finite) set of pairs $\langle A_i, x_i \rangle$, where: (i) $A_i$ is an atom over $\mathcal{R}_{MLN}$, $\mathcal{V}_{MLN}$, and $\Delta_{MLN}$; (ii) $x_i \in \{0, 1\}$; and (iii) for any two pairs $\langle A, x \rangle, \langle B, y \rangle \in \lambda$, there does not exist substitution $\theta$ that unifies $A$ and $B$. If $|\lambda| = |X|$ and all $\langle A, x_i \rangle \in \lambda$ are such that $A$ is ground, then $\lambda$ is called a *(possible) world*.

---

[1] http://research.cs.wisc.edu/hazy/tuffy/

Intuitively, a probabilistic annotation $\lambda$ is used to describe the class of events in which the random variables in an MLN are compatible with the settings of the random variables described by $\lambda$, i.e., each $X_i$ has the value $x_i$.

**Definition 2.** Let $F$ be the FOL translation of an $\mathcal{EL}^{++}$ axiom, and $\lambda$ be a probabilistic annotation; a *probabilistic $\mathcal{EL}^{++}$ axiom* is of the form $F : \lambda$. We also refer to probabilistic axioms as *annotated formulas*.

Essentially, probabilistic axioms hold whenever the events associated with their annotations occur. Note that whenever a random variable's value is left unspecified in an annotation, the variable is *unconstrained*; in particular, an empty annotation means that the formula holds in every possible world (we sometimes refer to these axioms as *crisp*).

**Definition 3.** Let $O$ be a set of (FOL translations of) probabilistic $\mathcal{EL}^{++}$ axioms and $M$ be an MLN. A *tightly coupled probabilistic $\mathcal{EL}^{++}$ ontology* (TCP ontology, or knowledge base) is of the form $KB = (O, M)$, where the probabilistic annotations of formulas in $O$ are relative to $M$.

Recall that random variables in our MLN setting are Boolean and written in the form of atoms over $\mathcal{R}_{MLN}$, $\mathcal{V}_{MLN}$, and $\Delta_{MLN}$; if $a$ is such an atom, $a = 1$ (resp., $a = 0$) denotes that the variable is *true* (resp., *false*); we also use the notation $a$ and $\neg a$, respectively.

**Definition 4.** Let $KB = (O, M)$ be a *probabilistic $\mathcal{EL}^{++}$ ontology*, and $\lambda$ be a possible world. The (non-probabilistic) $\mathcal{EL}^{++}$ ontology *induced* from $KB$ by $\lambda$, denoted $O_\lambda$, is the set $\{\theta_i F_i \mid F_i : \lambda_i \in O \text{ and } \theta_i \lambda_i \subseteq \lambda\}$, where $\theta_i$ is an mgu for $\lambda$ and $\lambda_i$.

The annotation of ontological axioms offers a clear modeling advantage by enabling a *clear separation of concerns* between the task of ontological modeling and the task of modeling the uncertainty around the axioms in the ontology. More precisely, in our formalism, it is possible to express the fact that the probabilistic nature of an ontological axiom is determined by elements that are *outside of the domain modeled by the ontology*.

**Example 3.** Consider the form-labeling ontology of Example 1 and the MLN of Example 2. In general, we expect only certain axioms to be probabilistic, while others are necessarily crisp. In our case, the fact that fields and text blocks are disjoint sets of objects, and that fields are labeled by text blocks are considered crisp axioms, since we do not want these assumptions to be violated by any model. However, we want to accept models where some field is left unlabeled, and therefore violating the axiom $Field \sqsubseteq \exists label.Text$ is possible. In addition, we want to link the probability that this axiom holds to the heuristics used to produce the actual labeling of the field, which are out of the domain of the form-labeling ontology. The following is a possible probabilistic $\mathcal{EL}^{++}$ ontology modeling this setup, where the first-order representation of the $\mathcal{EL}^{++}$ axioms discussed above is used to explicitly state the relationships between variables and constants in the MLN and in the ontology.

$\forall X.\textit{field}(X) \rightarrow \exists Y.\textit{label}(X, Y) \wedge \textit{text}(Y) :$
$$\{\langle \textit{canLabel}(Y, X), 1\rangle\};$$

$\forall X.\textit{label}(X, Y) \rightarrow \textit{field}(X) : \{\};$

$\forall X.\textit{label}(X, Y) \rightarrow \textit{text}(Y) : \{\};$

$\forall X.\textit{field}(X) \wedge \textit{text}(X) \rightarrow \bot : \{\}.$ ∎

*Data Complexity.* In this setting, we extend the usual concept of data complexity as follows: the set of formulas in the MLN are considered to be fixed, as is the TBox; on the other hand, the sets $\mathbf{I}$ and $\Delta_{MLN}$ are not, and therefore the ground ABox on the ontology side and the set of random variables on the MLN side *are not fixed*, either.

### 3.2 Semantics

The semantics of TCP ontologies is given relative to probabilistic distributions over *interpretations* of the form $\mathcal{I}_{MLN} = \langle D, w \rangle$, where $D$ is a database over $\mathbf{I} \cup \Delta_N$, and $w$ is a world. We usually abbreviate "$true : \lambda$" with "$\lambda$".

**Definition 5.** An interpretation $\mathcal{I}_{MLN} = \langle D, w \rangle$ *satisfies* an annotated formula $F : \lambda$, denoted $\mathcal{I}_{MLN} \models F : \lambda$, iff whenever there exists an mgu $\theta$ such that for all $\langle V_i, x_i \rangle \in \lambda$ it holds that $X_i = \theta V_i$ and $w[i] = x_i$, then $D \models \theta F$.

A *probabilistic interpretation* is then a probability distribution $Pr$ over the set of all possible interpretations such that only a finite number of interpretations are mapped to a non-zero value. The probability of an annotated formula $F : \lambda$, denoted $Pr(F : \lambda)$, is the sum of all $Pr(\mathcal{I}_{MLN})$ such that $\mathcal{I}_{MLN}$ satisfies $F : \lambda$.

**Definition 6.** Let $Pr$ be a probabilistic interpretation, and $F : \lambda$ be an annotated formula. We say that $Pr$ *satisfies* (or is a *model* of) $F : \lambda$ iff $Pr(F : \lambda) = 1$. Furthermore, $Pr$ is a model of a probabilistic $\mathcal{EL}^{++}$ ontology $KB = (O, M)$ iff: (i) $Pr$ satisfies all annotated formulas in $O$, and (ii) $1 - Pr(\textit{false} : \lambda) = Pr_M(\lambda)$ for all possible worlds $\lambda$, where $Pr_M(\lambda)$ is the probability of $\bigwedge_{\langle V_i, x_i \rangle \in \lambda}(V_i = x_i)$ in the MLN $M$ (and computed in the same way as $P(X = x)$ in Section 2.2).

In Definition 6 above, condition (ii) is stating that the probability values that $Pr$ assigns are in accordance with those of MLN $M$ (note that the equality in the definition implies that $Pr(\textit{true} : \lambda) = Pr_M(\lambda)$) and that they are adequately distributed (since $Pr(\textit{true} : \lambda) + Pr(\textit{false} : \lambda) = 1$).

In the following, we are interested in computing the probabilities of atoms in a TCP ontology, working towards ranking of entailed atoms based on their probabilities.

**Definition 7.** Let $KB = (O, M)$ be a TCP ontology, and $a$ be a ground atom that is constructed from predicates and individuals in $KB$. The *probability* of $a$ in $KB$, denoted $Pr^{KB}(a)$, is the infimum of $Pr(a : \{\})$ subject to all probabilistic interpretations $Pr$ such that $Pr \models KB$.

Intuitively, an atom has the probability that results from summing the probabilities of all possible worlds under which the induced ontology entails the atom.

### 3.3 Ranking of Atoms based on Probabilities

In this paper, we focus on queries requesting the ranking of atoms based on their probability values.

**Definition 8.** Let $KB = (O, M)$ be a TCP ontology; the *answer* to a *ranking query* $Q = rank(KB)$ is a tuple $ans(Q) = \langle a_1, \ldots, a_n \rangle$ such that $\{a_1, \ldots, a_n\}$ are all of the atomic consequences of $O_\lambda$ for any possible world $\lambda$, and $i < j \Rightarrow Pr^{KB}(a_i) \geq Pr^{KB}(a_j)$.

The straightforward approach to answering ranking queries is to compute the probabilities of all atoms inferred by the knowledge base; unfortunately, computing exact probabilities of atoms is intractable.

**Theorem 1.** *Let $KB = (O, M)$ be a TCP ontology. Computing $\Pr^{KB}(a)$ is #P-hard in the data complexity.*

In the next section, we explore how this negative result can be avoided by considering a special kind of MLN that allows us to compute *scores* instead of probability values. Such scores are closely related to probabilities, and allow us to rank answers with respect to actual probability values.

## 4 Tractably Answering Ranking Queries

In this section, we consider a *special case of MLNs* that proves to be useful towards more tractable methods to answer ranking queries.

### 4.1 Conjunctive MLNs and Equivalence Classes

We now introduce a simple class of MLNs:

**Definition 9.** A *conjunctive* Markov logic network (cMLN) is an MLN in which all formulas $(F, w)$ in the set are such that $F$ is a conjunction of atoms, and $w \in \mathbb{R}$.

Informally, a cMLN is an MLN in which formulas are restricted to conjunctions of atoms. This restriction allows us to define an equivalence relation over the set of worlds. Given a cMLN $M$ and its grounding $gr(M, \Delta_{MLN})$, we use the notation $Sat(\lambda, M)$ to denote the set of all ground formulas $F$ in $gr(M, \Delta_{MLN})$ that are satisfied by world $\lambda$. It is easy to see that $gr(M, \Delta_{MLN})$ can be computed in polynomial time in the data complexity; we therefore work with ground cMLNs in the rest of the paper.

**Definition 10.** Let $M$ be a ground cMLN, and $\lambda_1$ and $\lambda_2$ be possible worlds. We say that $\lambda_1$ and $\lambda_2$ are *equivalent* with respect to $M$, denoted $\lambda_1 \sim_M \lambda_2$, iff $Sat(\lambda_1, M) = Sat(\lambda_2, M)$.

Clearly, $\sim_M$ is an equivalence relation; we denote the equivalence classes induced by this relation with $C_1, \ldots, C_N$; note that, in general, there are $2^{|gr(M, \Delta_{MLN})|}$ such classes, making it intractable to inspect all of them.

**Example 4.** Consider the MLN in Example 2, and the grounding discussed above (cf. Figure 3). In this case, each of $\psi_1$ to $\psi_4$ has one possible grounding, which we call $f_1$ to $f_4$. There are therefore 16 equivalence classes in this case (each ground formula can be negated or not), which form a partition of the $2^8 = 256$ possible worlds. ∎

There are several properties of equivalence classes that we can leverage. First, equivalence classes are not guaranteed to be non-empty, but it is simple to check for this condition.

**Theorem 2.** *Let $M$ be a ground cMLN, and $C$ be a $\sim_M$-equivalence class. Deciding $C = \emptyset$ can be done in polynomial time.*

Furthermore, generating worlds for a given equivalence class is also tractable:

**Theorem 3.** *Let $M$ be a ground cMLN, and $C$ be a $\sim_M$-equivalence class. All elements in $C$ can be obtained in linear time with respect to the size of the output.*

*Proof sketch.* Let $C$ be described by formula *pos* $\wedge$ *neg*, where *pos* is the conjunction of all formulas from $M$ that are true in $C$, while *neg* is the conjunction of the negations of all formulas not true in $C$. The set of all atoms in the Herbrand base can be divided into two sets: *det* and *undet*; the former contains all (possibly negated) atoms that are determined by *pos*, while the latter contains the rest. The worlds in $C$ are obtainable by traversing *neg* and assigning truth values to atoms in this formula in all possible ways that make $C$ true, without contradicting the atoms in *det*. □

Finally, we point out an important characteristic of equivalence classes with respect to probability values:

**Theorem 4.** *Let $M$ be a ground cMLN, and $\lambda_1$ and $\lambda_2$ be possible worlds. If $\lambda_1 \sim_M \lambda_2$, then $\Pr_M(\lambda_1) = \Pr_M(\lambda_2)$.*

*Proof sketch.* Follows from Definition 10. Since both worlds belong to the same class, they satisfy (and do not satisfy) the same formulas, and so the value $\exp\left(\sum_j w_j \cdot n_j(\lambda_i)\right)$ from Equation 1 is the same for both worlds; the denominator (factor $Z$) is equal across all worlds, which means that their probabilities are equal. □

However, computing exact probabilities remains intractable in cMLNs.

**Theorem 5.** *Let $KB = (O, M)$ be a TCP ontology, where $M$ is a cMLN. Deciding $\Pr^{KB}(a) \geq k$ is PP-hard in the data complexity.*

The complexity class *PP* contains problems decidable by a probabilistic Turing machine in polynomial time, with error probability less than 1/2; like #*P*, a polynomial time Turing machine with a *PP* oracle can solve all problems in the polynomial hierarchy [21]. This negative result does not prevent us, however, from tractably *comparing* the probabilities of two worlds.

**Proposition 1.** *Let $M$ be a cMLN, and $\lambda_1$ and $\lambda_2$ be possible worlds. Deciding whether $\Pr_M(\lambda_1) \leq \Pr_M(\lambda_2)$ is in PTIME in the data complexity.*

The basic intuition behind this result is that we can compute the term $n_i(x)$ in Equation 1 in polynomial time; since the denominator in this equation is the same for all worlds, $\Pr_M(\lambda_1) \leq \Pr_M(\lambda_2)$ can be decided by only computing the numerators in this equation.

**Example 5.** Consider the following worlds relative to the MLN in Example 2 and the grounding from Example 4:

$$\begin{aligned}\lambda_1 &= \{\mathit{canLabel}(f,\ell), \mathit{hor}(\ell,f), \neg \mathit{left}(\ell,f), \mathit{adj}(\ell,f),\\ &\quad \mathit{right}(\ell,f), \neg \mathit{top}(\ell,f), \neg \mathit{nw}(\ell,f), \neg \mathit{ver}(\ell,f)\};\\ \lambda_2 &= \{\mathit{canLabel}(f,\ell), \neg \mathit{hor}(\ell,f), \neg \mathit{left}(\ell,f), \mathit{adj}(\ell,f),\\ &\quad \neg \mathit{right}(\ell,f), \mathit{top}(\ell,f), \neg \mathit{nw}(\ell,f), \mathit{ver}(\ell,f)\}.\end{aligned}$$

A quick inspection of the formulas in the MLN allows us to conclude that $\lambda_1 \models \neg f_1 \wedge f_2 \wedge \neg f_3 \wedge \neg f_4$, while $\lambda_2 \models \neg f_1 \wedge \neg f_2 \wedge f_3 \wedge \neg f_4$. Therefore, taking into account the weights of each formula, we can see that $\Pr(\lambda_1) > \Pr(\lambda_2)$; in fact, even though we cannot (tractably) compute the actual probabilities, we can conclude that $\Pr(\lambda_1) = \frac{e^6}{e^5} \Pr(\lambda_2)$, or that $\lambda_1$ is approximately 2.7 times more probable than $\lambda_1$. ∎

### 4.2 An Anytime Algorithm

The results in the previous section point the way towards Algorithm anytimeRank for answering ranking queries; the pseudocode for this algorithm is shown in Figure 4; in the rest of this section, we will discuss its properties.

Algorithm anytimeRank takes a probabilistic ontology in which the MLN is assumed to be a ground cMLN (recall that the grounding can be computed in polynomial time in the data complexity); the other input corresponds to a *stopping condition* that can be based on whatever the user considers important (time, number of steps, number of inspected worlds, etc); cf. Section 4.2.2 for a discussion on ways to define the stopping condition based on properties of the output offering correctness guarantees.

The main `while` loop in line 3 iterates through the set of equivalence classes relative to $M$. Subroutine *compMostProbEqClass*, invoked in line 4, computes the $i$-th most probable equivalence class. Note that this can simply be done by taking the formulas in $M$ and sorting them with respect to their weights; the classes are then generated by keeping track of a Boolean vector of which formulas are true and which are false. The next `while` loop, in line 7, is in charge of going through the current equivalence class. Subroutine *computePossWorld* takes the current class and a set of already inspected worlds and computes a new world (not in $S$). This can be done as described in the proof sketch of Theorem 3; in particular, the possible combinations of atoms in formula *neg* can be traversed in order, without the need to explicitly keep track of a set like $S$. The final lines of this loop take the computed world and obtain the atomic consequences from the (non-probabilistic) induced subontology (line 10), and adds the *score* to each such atom (lines 11 and 12). The *score* of a class consists of $e$ to the power of the sum of the weights of formulas that are true in that class. Line 13 updates the output set of atoms. Finally, set *out* is returned in decreasing order of score.

**Example 6.** Consider the probabilistic $\mathcal{EL}^{++}$ ontology $\Phi = (O, M)$, where $O$ is given as follows:

$$\begin{aligned}\forall X\, p(X) \to q(X) &\quad : \{\langle m(X), 1\rangle, \langle n(X), 0\rangle\};\\ p(a) &\quad : \{\langle m(a), 1\rangle\};\\ p(b) &\quad : \{\langle n(a), 1\rangle\};\\ p(c) &\quad : \{\langle m(c), 1\rangle, \langle n(c), 0\rangle\},\end{aligned}$$

and $M$ is given by $\{(m(X), 1.5), (n(X), 0.8)\}$. Suppose we ground $M$ with the set of constants $\{a, b, c\}$, yielding:

$$\begin{aligned}&f_1 : (m(a), 1.5), \quad f_2 : (n(a), 0.8), \quad f_3 : (m(b), 1.5),\\ &f_4 : (n(b), 0.8), \quad f_5 : (m(c), 1.5), \quad f_6 : (n(c), 0.8).\end{aligned}$$

This setup therefore yields $2^6 = 64$ equivalence classes (in this case, we have exactly one world per class). Figure 5 shows a subset of these classes in decreasing order of score (as they will be inspected by Algorithm anytimeRank). The algorithm will proceed as follows; cf. Figure 5 for the description of the classes (we use classes $C_i$ to denote the single world $\lambda_i$ in that class):

$O_{C_1} \models \{p(a), p(b)\}$; add $e^{6.9}$ to *score*($p(a)$) and *score*($p(b)$);

$O_{C_2} \models \{p(a), p(b), p(c), q(c)\}$; add $e^{6.1}$ to *score*($p(a)$), *score*($p(b)$), *score*($p(c)$), and *score*($q(c)$);

$O_{C_3} \models \{p(a)\}$; add $e^{6.1}$ to *score*($p(a)$);

$O_{C_4} \models \{p(a), p(c), q(c)\}$; add $e^{6.1}$ to *score*($p(a)$), *score*($p(c)$), and *score*($q(c)$);

$O_{C_5} \models \{p(a), p(b), q(a)\}$; add $e^{5.3}$ to *score*($p(a)$), *score*($p(b)$), and *score*($q(a)$).

If the algorithm is stopped at this point, the scores are as follows (approximate, and in descending order):

$p(a) : 2530.18, p(b) : 1638.47, p(c) : 891.71, q(c) : 891.71$ ∎

#### 4.2.1 Correctness and Running Time of anytimeRank

The correctness of this algorithm lies in the fact that if all classes are inspected, the returned output set is clearly the answer to the ranking query. In the general case, only a

```
Algorithm anytimeRank(KB = (O, M), stopCond)
// M is assumed to be a ground cMLN
 1. score:= empty mapping from atoms to ℝ (default 0);
 2. out:= empty set of atoms;   i := 1;
 3. while (i ≤ 2^|M|) and !stopCond do begin
    // i ranges over classes of possible worlds
 4.    C := compMostProbEqClass(M, i);
 5.    S := ∅;
 6.    i := i + 1;
 7.    while (|S| ≠ |C|) and !stopCond do
 8.       λ := computePossWorld(C, S);
          // compute world s.t. λ ∈ C and λ ∉ S
 9.       S := S ∪ {λ};
10.       O_λ := getInducedOnt(O, λ);
11.       for all atoms a ∈ atomicCons(O_λ) do
12.          score(a)+= exp (∑_{F_j ∈ M, C ⊨ F_j} w_j);
13.       out:= out ∪ atomicCons(O_λ);
14. end;
15. return out sorted in dec. order according to score.
```

Figure 4: An anytime algorithm to compute the answer to a ranking query over a TCP ontology (refer to text for description of subroutines).

| Class | $f_1$ | $f_3$ | $f_5$ | $f_2$ | $f_4$ | $f_6$ | $\sum_j w_j$ |
|-------|-------|-------|-------|-------|-------|-------|--------------|
| $C_1$ | 1 | 1 | 1 | 1 | 1 | 1 | 6.9 |
| $C_2$ | 1 | 1 | 1 | 1 | 1 | 0 | 6.1 |
| $C_3$ | 1 | 1 | 1 | 1 | 0 | 1 | 6.1 |
| $C_4$ | 1 | 1 | 1 | 0 | 1 | 1 | 6.1 |
| $C_5$ | 1 | 1 | 1 | 1 | 0 | 0 | 5.3 |
|       |   |   |   | ... |   |   |     |

Figure 5: Equivalence classes from Example 6, sorted in descending order of the *score* assigned to possible worlds that belong to them (e to the power of the value in the last column); only 5 of the 64 classes are shown here.

subset of the worlds will be inspected; since the probabilities of worlds in a given equivalence class are all equal (Theorem 4), and this value depends directly on the formulas in the cMLN that are satisfied by the class, the iteration through the equivalence classes in decreasing order of probability is the optimal path to take. Though the total running time of course depends on the stopping condition, Theorem 3, along with the way in which equivalence classes are manipulated (as described above), guarantee a running time that is polynomial in the data complexity as long as the combined number of inspected worlds and equivalence classes is bounded by a polynomial as well.

#### 4.2.2 Bounding the Error of anytimeRank

The following proposition provides a bound on the total "mass" of score that remains unassigned by our algorithm after a certain number of iterations.

**Proposition 2.** *Let $KB = (O, M)$ be a TCP ontology where $M$ is a ground cMLN with $n$ ground atoms, and let $C_1, ..., C_{2^{|M|}}$ be the set of equivalence classes of $M$ sorted in decreasing order of their score. Then, after analyzing s worlds and t classes with Algorithm anytimeRank, the total unassigned class score mass is bounded by above by $U = (2^n - s) \cdot \exp\left(\sum_{F_j \in M, C_{t+1} \models F_j} w_j\right)$.*

This result is useful, for instance, in determining a *provably correct partial order* over the output of the algorithm. For example, if the output is $\{(a, 120), (b, 90), (c, 80), (d, 10)\}$ and $U = 30$, we can safely conclude that $\Pr^{KB}(a) > \Pr^{KB}(c)$, $\Pr^{KB}(a) > \Pr^{KB}(d)$, $\Pr^{KB}(b) > \Pr^{KB}(d)$, and $\Pr^{KB}(c) > \Pr^{KB}(d)$.

**Theorem 6.** *Let out be the output of Algorithm anytimeRank and $U$ be the bound on the unassigned score mass as computed in Proposition 2. The partial order $\leq_U$ defined as: $a \leq_U b$ iff $s_a + U \leq s_b$, where $(a, s_a), (b, s_b) \in out$, is such that if $a \leq_U b$ then $\Pr(a) \leq \Pr(b)$.*

Therefore, Theorem 6 allows us to glimpse into the total order over the set of atoms as established by the true probability values, without actually computing them.

## 5 Related Work

Ontology languages, rule-based systems, and their integrations are central for the Semantic Web [2]. Although many approaches exist to tight, loose, or hybrid integrations of ontology languages and rule-based systems, to our knowledge there is very little work on the combination of tractable description logics with MLNs. Probabilistic ontology languages in the literature can be classified according to the underlying ontology language, the supported forms of probabilistic knowledge, and the underlying probabilistic semantics (see [14] for a recent survey). Some early approaches [11] generalize the description logic $\mathcal{ALC}$ and are based on propositional probabilistic logics, while others [12] generalize the tractable DL CLASSIC and $\mathcal{FL}$, and are based on Bayesian networks as underlying probabilistic semantics. The fairly recent approach in [13], generalizing the expressive DL $\mathcal{SHIF}(\mathbf{D})$ and $\mathcal{SHOIN}(\mathbf{D})$ behind the sublanguages *OWL Lite* and *OWL DL*, respectively, of the Web ontology language *OWL* [18], is based on probabilistic default logics, and allows for rich probabilistic terminological and assertional knowledge. Other recent approaches [23] generalize OWL by probabilistic uncertainty using Bayesian networks.

In the probabilistic description logics literature, the most closely related work is that of Prob-$\mathcal{EL}$ [15, 10], a probabilistic extension to $\mathcal{EL}$ that belongs to a family of probabilistic DLs derived in a principled way from Halpern's probabilistic first-order logic [4]. One limitation in this line of research is that probabilistic annotations are somewhat restricted, leading to the inability to express uncertainty about certain kinds of general knowledge; note that our formalism does not suffer from this drawback, since probabilistic annotations can be associated with any axiom. In [15], the authors study various logics in this family and show complexity of reasoning, ranging from PTIME for weak variants of $\mathcal{EL}$ to undecidable for expressive variants of $\mathcal{ALC}$. Prob-$\mathcal{EL}$ is more closely studied in [10], where the authors show that reasoning is PTIME as long as (i) probability values are restricted to 0 and 1, and (ii) probabilistic annotations are only allowed on concepts; if (i) is dropped, then it becomes EXPTIME-complete, while if (ii) is dropped it becomes PSPACE-hard. The complexity results in our work are further testament to how intractable simple tasks become when probabilistic computations are involved, even when the starting point is a tractable logic.

Other recent efforts focused on extending $\mathcal{EL}$ DLs with probabilistic uncertainty include [5], [9], and [17]. In [5], a formalism is presented in which probability assessments are only allowed on ABoxes. The authors study the problem of satisfiability of KBs, which involves determining whether there exists a probability distribution that satisfies all the assignments over the ABoxes. The work of [9] is similar in spirit to [5], but there an MLN is employed in order to infer the probabilities of atoms in the ABox as a means to generate explanations as part of abductive inference. Though they use MLNs, this work is quite different from ours since only the ABox is assumed to be probabilistic, and the assertions are themselves part of the MLN instead of being annotated by external events as in our formalism. In [17], the authors present ELOG, which is $\mathcal{EL}^{++}$ without nominals or concrete domains combined with probabilistic log-linear models (a class which contains MLNs). The resulting probabilistic formalism basically assigns weights to axioms reflecting how likely the axiom is to hold. The main problem then corresponds to finding the most probable coherent ontology, a problem that is essentially different from the one tackled here.

Finally, a related formalism from the recent databases literature is that of probabilistic Datalog+/– [8, 7]. Datalog+/– is a language that arose from the generalization of rule-based constraints with the goal of expressing ontological axioms. The probabilistic extension in [8, 7] also makes use of MLNs, though the integration is loose in the sense that probabilistic annotations cannot refer to objects in the ontology, which leads to data-tractable algorithms but also limits the expressive power of the formalism.

## 6 Discussion, Conclusions, and Future Work

In this work, we have extended the DL $\mathcal{EL}^{++}$ with probabilistic uncertainty, based on the annotation of axioms. Such annotations refer to events whose probabilities are described by an associated MLN; one of the advantages of this formalism is that it is tightly coupled, which means that probabilistic annotations can refer to objects in the ontology. The proposed application of our formalism is in managing uncertainty in the Semantic Web, showing examples of how it can be applied in the analysis of Web forms, an important task in information extraction efforts.

Our focus here is on ranking queries, which request the set of atomic inferences sorted in descending order of probability. The algorithm we developed works in an anytime fashion, and therefore allows partial computations depending on the available resources; most importantly, we provide bounds on the error incurred by runs of this algorithm and conditions that allow to conclude when certain pairs in the output are correctly ordered. This algorithm works over cMLNs, in which only conjunctions of atoms are allowed. Regarding the expressivity of cMLNs, we can say that: (i) They are rich enough to simulate disjunction for a specific propositional subset. For instance, if the formula $p(X) \lor q(X)$ with a given weight needs to be enforced for the subset of individuals $\{a, b\}$, MLN learning algorithms can be directed to give corresponding weights to the specific worlds in which $\big(p(a)$ or $q(a)\big)$ and $\big(p(b)$ or $q(b)\big)$ hold. So, it is possible to represent certain special cases that may need to be handled. (ii) For the case of (atomic) negation, it is known to be representable via negative weights. (iii) Though material implications cannot be represented, this sort of constraint is more adequately placed on the ontology side, and the TBox is capable of representing them.

Regarding the practical applicability of the formalism, we can say that, despite probabilistic instance-checking being already intractable for cMLNs, it is often possible to bound the number of scenarios in practice. Consider, for instance, the problem of reasoning over data structures on the Web (related to the running example). Web pages are usually processed in isolation, since structured data do not usually span across pages and are often delimited by a certain DOM sub-tree. This implies that the number of possible worlds is bounded by a function of the constants appearing in a subset of the page. This bound can work together with the good computational behavior of cMLNs to allow tractability of reasoning in practice. Other applications where cMLNs can be leveraged are semantic and natural language-based Web search.

Future work involves investigating other DLs that can be extended in this manner, and pushing the known line between tractability and expressivity. We also need to empirically evaluate our approach both on synthetic and real-world data, as well as studying the application of other techniques such as random sampling, which may provide increased scalability at the cost of lost guarantees.

**Acknowledgments**

This work was supported by the Engineering and Physical Sciences Research Council (EPSRC) grant EP/J008346/1 ("PrOQAW"), the Oxford Martin School grant LC0910-019, the European Research Council (ERC) grant FP7/246858 ("DIADEM"), a Google Research Award, and a Yahoo! Research Fellowship.

# References

[1] BAADER, F., BRANDT, S., AND LUTZ, C. Pushing the el envelope further. In *Proc. of OWLED* (2008), K. Clark and P. F. Patel-Schneider, Eds.

[2] BERNERS-LEE, T., HENDLER, J., AND LASSILA, O. The Semantic Web. *Scientific American 284* (2002), 34–43.

[3] CALÌ, A., GOTTLOB, G., AND LUKASIEWICZ, T. A general Datalog-based framework for tractable query answering over ontologies. In *Journal of Web Semantics* (2012).

[4] FAGIN, R., HALPERN, J. Y., AND MEGIDDO, N. A logic for reasoning about probabilities. *Information and Computation 87*, 1/2 (1990), 78–128.

[5] FINGER, M., WASSERMANN, R., AND COZMAN, F. G. Satisfiability in EL with sets of probabilistic aboxes. In *Proc. of DL* (2011).

[6] FURCHE, T., GOTTLOB, G., GRASSO, G., GUO, X., ORSI, G., AND SCHALLHART, C. OPAL: Automated form understanding for the deep Web. In *Proc. of WWW* (2012).

[7] GOTTLOB, G., LUKASIEWICZ, T., AND SIMARI, G. I. Answering threshold queries in probabilistic Datalog+/– ontologies. In *Proc. of SUM* (2011), LNCS, pp. 401–414.

[8] GOTTLOB, G., LUKASIEWICZ, T., AND SIMARI, G. I. Conjunctive query answering in probabilistic Datalog+/– ontologies. In *Proc. of RR* (2011), LNCS, pp. 77–92.

[9] GRIES, O., MÖLLER, R., NAFISSI, A., ROSENFELD, M., SOKOLSKI, K., AND WESSEL, M. A probabilistic abduction engine for media interpretation based on ontologies. In *Proc. of RR* (2010), pp. 182–194.

[10] GUTIÉRREZ-BASULTO, V., JUNG, J. C., LUTZ, C., AND SCHRÖDER, L. A closer look at the probabilistic description logic Prob-EL. In *Proc. of AAAI* (2011).

[11] HEINSOHN, J. Probabilistic description logics. In *Proc. of UAI* (1994), pp. 311–318.

[12] KOLLER, D., LEVY, A., AND PFEFFER, A. P-CLASSIC: A tractable probabilistic description logic. In *Proc. of AAAI* (1997), pp. 390–397.

[13] LUKASIEWICZ, T. Expressive probabilistic description logics. *Artificial Intelligence 172* (2008), 852–883.

[14] LUKASIEWICZ, T., AND STRACCIA, U. Managing uncertainty and vagueness in description logics for the semantic web. *Journal of Web Semantics 6* (2008), 291–308.

[15] LUTZ, C., AND SCHRÖDER, L. Probabilistic description logics for subjective uncertainty. In *Proc. of KR* (2010), pp. 393–403.

[16] MADHAVAN, J., KO, D., KOT, L., GANAPATHY, V., RASMUSSEN, A., AND HALEVY, A. Google's deep Web crawl. In *Proc. of VLDB* (2008), pp. 1241–1252.

[17] NOESSNER, J., AND NIEPERT, M. ELOG: a probabilistic reasoner for OWL EL. In *Proc. of RR* (2011), pp. 281–286.

[18] PATEL-SCHNEIDER, P. F., HAYES, P., AND HORROCKS, I. OWL Web Ontology Language semantics and abstract syntax. W3C Recommendation, 2004. Available at http://www.w3.org/TR/owl-semantics/.

[19] PEARL, J. *Probabilistic reasoning in intelligent systems: networks of plausible inference.* 1988.

[20] RICHARDSON, M., AND DOMINGOS, P. Markov logic networks. *Machine Learning 62* (2006), 107–136.

[21] TODA, S. On the computational power of PP and $\oplus$P. In *Proc. of FOCS* (1989), pp. 514–519.

[22] WWW CONSORTIUM. Uncertainty Reasoning for the World Wide Web – W3C Incubator Group Report. Available at: http://www.w3.org/2005/Incubator/urw3/XGR-urw3/.

[23] YANG, Y., AND CALMET, J. OntoBayes: An ontology-driven uncertainty model. In *Proc. of IAWTIC* (2005), pp. 457–463.